\pdfoutput=1

\documentclass[11pt]{article}

\usepackage[final]{acl}

\usepackage{times}
\usepackage{latexsym}
\usepackage{rotating}
\usepackage[T1]{fontenc}

\usepackage{microtype}

\usepackage{inconsolata}

\usepackage{graphicx}

\usepackage{hyperref} 
\usepackage{amsmath}
\usepackage{enumitem}
\usepackage{tcolorbox}

\usepackage{xcolor}
\usepackage{ifthen}
\usepackage[normalem]{ulem}
\usepackage{graphicx}
\usepackage{booktabs}
\usepackage{amsmath}
\usepackage{amssymb}

\usepackage[utf8]{inputenc} 
\usepackage{graphicx}
\usepackage{booktabs}
\usepackage{array}
\usepackage{rotating} 
\usepackage{CJKutf8}

\definecolor{myorange}{HTML}{FEAE03}
\definecolor{myturquois}{HTML}{01AB8F}
\definecolor{mypink}{HTML}{D31876}

\definecolor{brightred}{HTML}{E55347} 
\definecolor{orange}{HTML}{FF8C00} 
\definecolor{yellowgreen}{HTML}{6B8E23} 
\definecolor{green}{HTML}{228B22} 

\usepackage{amsmath,amssymb}
 \usepackage{algorithm}
\usepackage{algpseudocode}

\usepackage{booktabs}
\usepackage{tabularx}
\usepackage{xcolor}

\newcommand{\improve}[1]{\textcolor{blue!70!black}{\scriptsize(+#1)}}

\newcommand{\improved}[1]{\textcolor{red!70!black}{\scriptsize(-#1)}}

\newtcolorbox{bluebox}[1][]{
	float,
  	title=#1,
	colback=myturquois!4,
	colframe=myturquois,
        top=1pt,           
        bottom=1pt,        
        left=0pt,          
        right=0pt,          
        before skip=0.65em, after skip=0.75em,
}

\title{Why Supervised Fine-Tuning Fails to Learn: A Systematic Study of Incomplete Learning in Large Language Models}

\author{
    Chao Xue\textsuperscript{1,$\star$},
    Yao Wang\textsuperscript{1,$\star$},
    Mengqiao Liu\textsuperscript{2},
    Di Liang\textsuperscript{2,3,$\dagger$}, \\
    Xingsheng Han\textsuperscript{2},
    Peiyang Liu\textsuperscript{5},
    Xianjie Wu\textsuperscript{2}, 
    Chenyao Lu\textsuperscript{2}, 
    Lei Jiang\textsuperscript{2},\\
    Yu Lu\textsuperscript{2},
    Haibo Shi\textsuperscript{2,3},
    Shuang Liang\textsuperscript{4},
    Minlong Peng\textsuperscript{2}, 
    Flora D. Salim\textsuperscript{1,$\dagger$} \\\\
    \textsuperscript{1} University of New South Wales, Australia, 
    \textsuperscript{2} Tencent Hunyuan, China,\\
    \textsuperscript{3} Tencent Yuanbao, China,
    \textsuperscript{4} UESTC, China ,
    \textsuperscript{5} Peking University, China \\\\
    \texttt{xuechao8071@gmail.com; flora.salim@unsw.edu.au}
}

\begin{document}
\maketitle

\footnotetext[1]{$\star$ Equal Contribution.$\dagger$ Corresponding Author.}
\footnotetext[2]{This work was completed by Xue Chao and Yao Wang under Di Liang’s supervision.}

\begin{abstract}
Supervised Fine-Tuning (SFT) is the standard approach for adapting large language models (LLMs) to downstream tasks. However, we observe a persistent failure mode: even after convergence, models often fail to correctly reproduce a subset of their own supervised training data. We refer to this behavior as the \textbf{\emph{Incomplete Learning Phenomenon} (ILP)}.
This paper presents the first systematic study of ILP in LLM fine-tuning. We formalize ILP as post-training failure to internalize supervised instances and demonstrate its prevalence across multiple model families, domains, and datasets. Through controlled analyses, we identify five recurrent sources of incomplete learning: (1) missing prerequisite knowledge in the pre-trained model, (2) conflicts between SFT supervision and pre-training knowledge, (3) internal inconsistencies within SFT data, (4) left-side forgetting during sequential fine-tuning, and (5) insufficient optimization for rare or complex patterns.
We introduce a diagnostic-first framework that maps unlearned samples to these causes using observable training and inference signals, and study several targeted mitigation strategies as causal interventions. Experiments on Qwen, LLaMA, and OLMo2 show that incomplete learning is widespread and heterogeneous, and that improvements in aggregate metrics can mask persistent unlearned subsets. The findings highlight the need for fine-grained diagnosis of what supervised fine-tuning fails to learn, and why. 
\end{abstract}

\section{Introduction}

Supervised Fine-Tuning has become the dominant paradigm for adapting large language models (LLMs) to downstream applications such as question answering, dialogue generation, and domain-specific reasoning \cite{hou2024raw,zhao2024supervised}. By leveraging relatively small but carefully curated labeled datasets, SFT enables pre-trained models to align their behavior with task-specific objectives while retaining general linguistic competence. As a result, SFT is widely regarded as a reliable and efficient mechanism for specialization.

\begin{figure}
\centering
\includegraphics[width=0.49\textwidth]{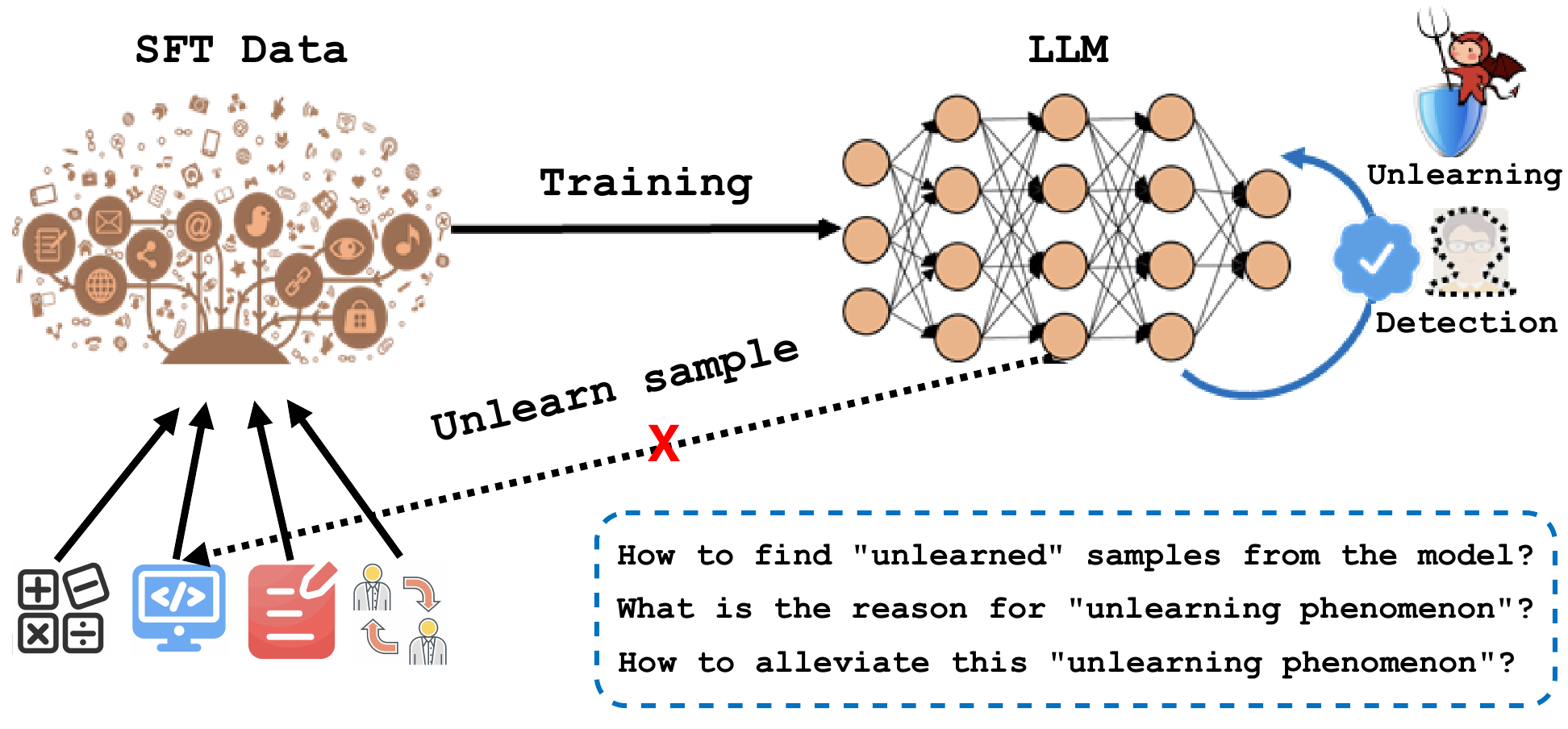}
\caption{\label{fig:example}Schematic illustration of the incomplete learning phenomenon, where testing the model on the initial training set after fine-tuning reveals that certain samples or patterns were not effectively learned during SFT.}
\end{figure}

Despite its widespread adoption, SFT exhibits a subtle but consequential failure mode that is insufficiently understood. In practice, we observe that even after training loss convergence and extensive hyperparameter tuning, LLMs frequently fail to correctly answer a subset of their supervised training examples. These failures occur on the SFT dataset itself, rather than on held-out or out-of-distribution data, and persist across random seeds and evaluation settings. We refer to this behavior as the \emph{Incomplete Learning Phenomenon} (ILP).
Figure~\ref{fig:example} illustrates ILP: after fine-tuning, re-evaluating the model on its supervised training set reveals that certain instances or patterns remain consistently mispredicted. Importantly, ILP is distinct from catastrophic forgetting \cite{mccloskey1989catastrophic}, which concerns the loss of previously acquired capabilities, and from machine unlearning \cite{cao2015towards}, which is intentional. Instead, ILP reflects a failure to acquire or internalize parts of the supervision signal during SFT.

Understanding ILP is practically important for several reasons. First, SFT datasets, especially in expert domains such as law and medicine, are costly to construct, and incomplete learning directly reduces their utility. Second, unlearned samples are often not random; they tend to correspond to rare cases, compositional patterns, or knowledge-intensive instances, which disproportionately affect robustness and reliability. Third, aggregate evaluation metrics can obscure ILP: improvements on standard benchmarks may coexist with persistent failures on specific supervised instances.
Prior work has investigated challenges related to fine-tuning stability, data quality, and optimization dynamics \cite{gururangan2020dontstoppretrainingadapt,zhang2024dissecting,bengio2009curriculum,wang2026rethinking}. However, these studies typically focus on improving overall task performance rather than explaining \underline{\emph{which supervised knowledge fails to be learned a-}} \underline{\emph{nd why}}. As a result, existing approaches provide limited tools for diagnosing fine-tuning failures at the level of individual samples or patterns.

In this paper, we take a phenomenon-driven perspective. Our goal is not to propose a new fine-tuning algorithm, but to systematically characterize, diagnose, and validate the sources of incomplete learning in SFT. Through extensive empirical analysis, we identify five recurring contributors to ILP:
\textbf{(i)} Pre-training Knowledge Limitations, where the base model lacks prerequisite concepts needed to absorb the supervised signal;
\textbf{(ii)} Knowledge Conflicts, where SFT supervision contradicts entrenched pre-training knowledge;
\textbf{(iii)} Internal SFT Data Conflicts, arising from noisy or inconsistent annotations;
\textbf{(iv)} Left-Side Forgetting, where earlier supervised instances are overwritten during sequential fine-tuning;
\textbf{(v)} Insufficient Optimization for Complex Patterns, where rare or compositional structures receive inadequate training signal.

To operationalize this analysis, we introduce a diagnostic framework that associates unlearned samples with these causes using observable training and inference indicators, such as prediction consistency, entropy dynamics, and replay sensitivity. We further examine several targeted mitigation strategies, including continued pre-training, conflict-aware scheduling, and replay-based resampling, not as universally optimal solutions, but as controlled interventions to test the plausibility of each hypothesized cause.
We evaluate our framework on multiple LLMs (Qwen, LLaMA, and OLMo2) across diverse domains and tasks. The results demonstrate that incomplete learning is both prevalent and heterogeneous: no single intervention resolves all failures, and improvements in aggregate metrics can mask persistent unlearned subsets.

Overall, this work makes three contributions. First, it identifies and formalizes the Incomplete Learning Phenomenon as a measurable and reproducible failure mode in supervised fine-tuning. Second, it provides a systematic taxonomy and diagnostic framework that links unlearned supervised instances to distinct underlying causes. Third, it empirically shows that different sources of incomplete learning require different remedies, highlighting the limitations of one-size-fits-all fine-tuning strategies. Together, these findings argue for a shift from performance-centric evaluation of SFT toward fine-grained, learning-centric diagnosis, offering a foundation for more reliable and interpretable adaptation of large language models.


\begin{figure*}
\centering
\includegraphics[width=1.0\textwidth]{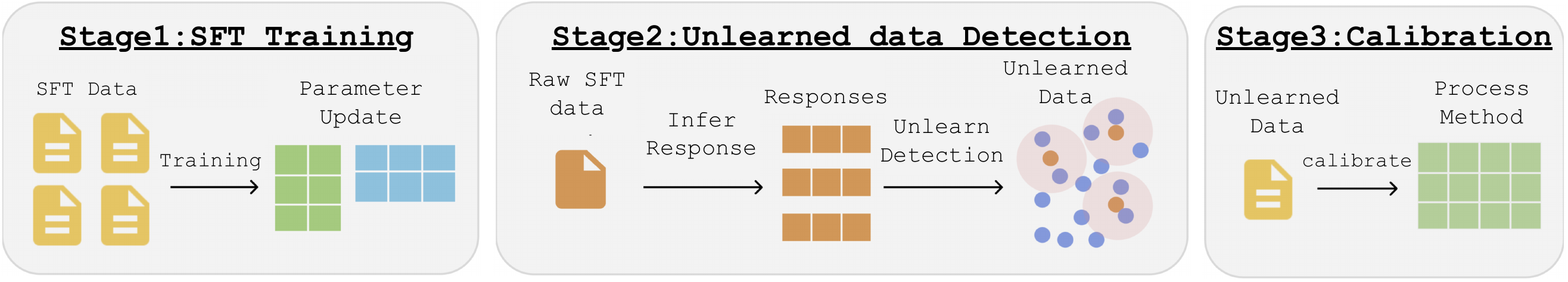}
\caption{\label{fig:frame}
The Incomplete Learning framework consists of three stages: (1) fine-tune on SFT data; (2) detect unlearned samples via re-evaluation; (3) calibrate model and data to fix them.}
\end{figure*}

\section{Related Works}
Recent research on large language models spans multiple directions:  stepwise distillation~\cite{chen2025improving,jiang2025drp,zhang2025find}, multi-hop temporal knowledge reasoning~\cite{wen2026reinforcement,xue2024question}, and security and robustness through jailbreak detection~\cite{hua2025rethinking} and backdoor analysis in reward learning~\cite{guo2026backdoorsrlvrjailbreakbackdoors}; structured representation learning for contextual semantic matching~\cite{xue2025structcoh} and empathetic dialogue modeling~\cite{ji2026strideedstrategygroundedstepwisereasoning}; multimodal referential understanding~\cite{wang2026one}; memorization-constrained story reasoning~\cite{jiang2026beyond}; and broader applications in AI governance~\cite{chen2026testing,chen2026beyond} and predictive analytics~\cite{hu2026predictive}.  
A common thread across many of these approaches is the reliance on high-quality, human- or model-generated reasoning demonstrations—typically injected into the model via \textit{supervised fine-tuning} (SFT)—to align behavior with desired reasoning patterns.

\paragraph{Supervised Fine-Tuning of Large Language Models}

LLMs show remarkable zero-shot capabilities \cite{brown2020language,wu2021yuan,hou2024large,song2023large}, leading to extensive efforts in enhancing their applicability via Supervised Fine-Tuning. To unlock their full potential, LLMs are often subjected to the SFT phase, which refines their ability to perform specific tasks and better align with human instructions \cite{ponti2023fine_tuning,li2026safety,gao2026decorl,huang2026semanticspaceexplorationexploitationrlvr}. This study broadens the traditional scope of SFT to incorporate diverse forms of sequence-to-sequence fine-tuning, including fine-tuning for human alignment, instruction adherence, and domain-specific task optimization \citep{zhou2023instructionfollowing,yuan2023rrhf,cheng2023m,zhang2024instruction,liu2026dpi}.
Recent research has explored multi-task instruction fine-tuning for pre-trained LLMs, aimed at enhancing their zero-shot performance across a broad range of downstream NLP tasks \citep{sanh2022multitask, khashabi2020unifiedqa}. Prominent efforts such as FLAN, which curated large-scale instruction datasets, have shown that models fine-tuned with such data \cite{chung2022scaling,singhal2022large} achieve improved zero-shot generalization. 
Although the generalization capabilities of LLMs in out-of-distribution domains have been extensively studied \cite{liu2024good,yuan2024revisiting,wang2024multiperspective}, the effect of multi-task fine-tuning on in-domain performance, and potential SFT-induced degradation of foundational abilities \cite{mukhoti2023fine,liu2025structural} or catastrophic forgetting \cite{kotha2023understanding}, remain critical areas of investigation. These challenges highlight the complexities our work on ILP addresses by focusing on why SFT data itself is not fully learned. With the rise of proprietary models like ChatGPT, the focus on SFT for better aligning LLMs with human intent has grown \citep{ouyang2022training}. Beyond crowd-sourcing, user logs \citep{vicuna2023,openchat} and LLM-assisted self-generated data \citep{wang2023selfinstruct,alpaca,cheng2023accelerating,lei2023instructerc,xu2023wizardlm,xue2023occuquest,wu2025progressive,mukherjee2023orca,wang2025not,wu2025tablebench} are increasingly used for SFT. Moreover, methods to improve the quality of SFT datasets have been proposed to enhance alignment with human preferences \citep{lima,tulu,instag,wu2025unleashing,liu2024makes,cui2023ultrafeedback}.
SFT has also proven valuable for domain-specific applications, excelling in areas such as mathematical reasoning \citep{gsm8k,rft, mammoth,gou2024tora,yue2024mammoth2,dai2025hope} and code generation tasks \citep{codealpaca,wizardcoder,wei2023magicoder,wu2025breaking}. Additionally, supervised fine-tuned LLMs have been leveraged to enhance interactivity by composing external commands, enabling the execution of a variety of highly complex downstream applications, such as tool integration \cite{yao2023react,yao2023tree,song2024knowledge,fu2024preact,liu2025stole,guo2026e3tirenhancedexperienceexploitation}. 

\paragraph{Data Quality and Multi-Stage Fine-Tuning}

Improving data quality is a recurring focal point in the SFT pipeline \cite{mazumder2023dataperf,li2024comateformer,liu2024resolving,li2024superfiltering}. Techniques such as data augmentation \cite{shorten2019survey,li2024local,fu2021fast} and active learning \cite{settles2009active} aim to enhance the diversity or informativeness of training examples. Prompt engineering \cite{lester2021power,liu2023local,liu2023time} has been introduced to reshape the input space, thereby encouraging more consistent model outputs. Additionally, knowledge distillation \cite{hinton2015distilling} is employed to transfer knowledge from larger teacher models to smaller or specialized student models, ensuring knowledge preservation while reducing model size or computational overhead \cite{sanh2019distilbert,liang2019adaptive,wang2022dabert,song2022improving,xue2023dual,chen2026sparse,gui2018transferring,zheng2022robust,liang2019asynchronous,hu2025joint,xue2026reasonneededefficientgenerative}.
Curriculum learning \cite{bengio2009curriculum,qianadaptive} arranges training samples in an order of increasing complexity, enabling models to develop foundational competencies before tackling more difficult examples. Such methods have demonstrated improved convergence rates and robustness \cite{platanios2019competence}. 
Multi-Task and multi-pass fine-tuning extend ideas by exposing the model to multiple related tasks or multi-step schedules, where earlier tasks are revisited \cite{dong2023abilities, ruder2017overview,ma2022searching,fei2022cqg}. These strategies highlight how training order, data scheduling, and repeated re-exposure to previously learned samples can reduce overfitting, mitigate forgetting, and improve generalization \cite{parisi2019continual}.

\paragraph{Scaling Laws in Large Language Models}

The remarkable performance of LLMs is driven by scaling model sizes, dataset volumes, and computational resources to unprecedented levels \cite{kaplan2020scaling}. Analyzing how performance across an exponential range of scales has become crucial. Research has explored scaling laws in pre-training \citep{palm2,hoffmann2022training}, transfer learning \citep{chronopoulou2019embarrassingly}, preference modeling \citep{gao2022rmscaling}, and mathematical reasoning \citep{rft}, underscoring the pivotal role of scaling in enhancing LLMs' capability.

\section{Methods}
Our method is designed to systematically diagnose and mitigate \emph{incomplete learning} phenomena in supervised fine-tuning of large language models. As illustrated in Figure~\ref{fig:frame}, the framework consists of two tightly coupled components: \textbf{Unlearned Sample Detection} and \textbf{Unlearned Sample Processing}.
The Unlearned Sample Detection module aims to identify training instances that are not effectively internalized by the model during SFT. Unlike conventional data filtering approaches that rely on static heuristics or annotation quality, we focus on samples that remain persistently mispredicted or unstable across training, indicating a failure of learning rather than noise. These unlearned samples form hidden bottlenecks that limit performance gains from additional data or training iterations.
Building on the detected unlearned samples, the Unlearned Sample Processing module analyzes their underlying characteristics and failure modes. Through empirical analysis, we categorize typical unlearned samples into five representative error types, each associated with distinct learning deficiencies. For each type, we design targeted processing strategies that directly address its root cause, rather than uniformly reweighting or discarding data. 

\subsection{Unlearned Sample Detection}
\label{subsec:detection}

A prerequisite for studying incomplete learning is a reliable mechanism to identify which supervised instances are not effectively learned after fine-tuning. In this work, we treat unlearned sample detection as a post-training measurement problem rather than an optimization objective. Specifically, we ask whether a model, after supervised fine-tuning (SFT) convergence, can consistently reproduce the supervision signal it has already seen.

\subsubsection{Sample-Level Evaluation}

SFT datasets typically consist of free-form text responses, which makes instance-level correctness difficult to assess in a standardized manner. To enable consistent measurement across heterogeneous datasets and tasks, we operationalize supervised responses into a multiple-choice (MC) format. This conversion is not intended to change the supervision content, but to provide a discrete and comparable evaluation interface.
Concretely, for each SFT instance, the original response is preserved as the correct option, while several semantically plausible but incorrect alternatives are constructed as distractors. The model is then required to select the correct option among a fixed set of candidates. Figure~\ref{fig:frame} illustrates the overall framework, and an example of this conversion is shown below.
\begin{bluebox}[Dataset Conversion Example]
\textbf{Original Sample.}
\underline{\textit{Prompt:}} Explain what deep learning is. \\
\underline{\textit{Response:}} Deep learning is a machine learning method based on neural networks.

\textbf{Converted Evaluation Form.}
\underline{\textit{Options:}} \\
A) is a machine learning method based on neural networks. \\
B) is a rule-based system. \\
C) is a reinforcement learning method. \\
D) is a traditional statistical method.
\end{bluebox}

The index of the correct option is recorded and used for subsequent evaluation. Importantly, this conversion is applied \emph{only for detection and analysis} and does not alter the original SFT training objective.

\subsubsection{Post-SFT Consistency Evaluation}

As shown in Figure~\ref{fig:frame}, unlearned sample detection is performed after the SFT process has converged. During fine-tuning, we monitor the training loss to ensure stable optimization and exclude under-training artifacts. After convergence, the entire SFT dataset is re-evaluated by the fine-tuned model using the MC-based interface.
For a dataset of $N$ supervised instances, we define sample-level correctness by whether the model selects the ground-truth option. The training-set accuracy is:
\begin{equation}
\label{eq:accuracy}
\mathrm{Acc} = \frac{1}{N} \sum_{i=1}^{N} \mathbb{I}\left(\arg\max_k \hat{y}_{i,k} = y_i\right),
\end{equation}



where $\hat{y}_{i,k}$ is predicted probability for option $k$ of instance $i$, and $y_i$ is the correct index. Accuracy is coarse and misses partial or unstable learning; thus, we use repeated sampling to reduce stochasticity.

\subsubsection{Robust Detection}

For each instance, we perform $N$ independent inference runs and compute its \emph{pass@N} rate, defined as the fraction of runs in which the model predicts the correct option. This metric reflects the consistency with which the supervision signal is recovered. In addition, we adopt a Best-of-$N$ (BoN) criterion, which selects the prediction with the highest confidence score among $N$ samples, providing a complementary upper-bound estimate of model capability.
An instance is considered \emph{unlearned} if its pass@N rate falls below a predefined threshold $T$. In our experiments, we set $T=0.2$ under BoN-5 sampling unless otherwise stated. We empirically verify that the identified unlearned instances are stable across random seeds and sampling runs, indicating that they are not artifacts of stochastic decoding.

\subsubsection{Empirical Prevalence of Unlearned Samples}

Applying this detection protocol across ten benchmark SFT datasets, we find that incomplete learning is widespread. On average, $15.3\% \pm 2.1\%$ of supervised instances remain unlearned after SFT convergence. This observation holds across model families and domains, suggesting that ILP is not an isolated or dataset-specific phenomenon.
For subsequent analysis, we construct a candidate set by selecting instances with pass@5 rates below the threshold under repeated BoN-5 sampling. From this set, we select the top-$K$ most severe cases based on error consistency, with $K=1000$ in our main experiments. These instances form the basis for fine-grained diagnosis in the following section.

\subsubsection{Knowledge-State Probing for Diagnostic Preparation}

To enable attribution of unlearned samples to potential causes, we probe the knowledge state of the base model prior to fine-tuning. For each candidate instance $x$, we first test whether the base model can correctly answer it in a zero-shot setting. We define a binary indicator of knowledge existence as
\begin{equation}
\label{eq:existence}
\mathcal{P}_{\mathrm{exist}}(x) = \mathbb{I}\left(\mathrm{Acc}(\mathcal{M}_{\mathrm{base}}(x)) > 0.8\right).
\end{equation}
In addition, we measure how the model's predictive distribution changes after SFT by computing the Jensen--Shannon divergence between the base and fine-tuned models,as:
\begin{equation}
\begin{aligned}
D_\text{JS}(P_\text{base} \| P_\text{SFT}) = \frac{1}{2}D_\text{KL}\left(P_\text{base} \| M\right) \\
+ \frac{1}{2}D_\text{KL}\left(P_\text{SFT} \| M\right)
\end{aligned}
\label{eq:consistency}
\end{equation}
where $M = (P_{\mathrm{base}} + P_{\mathrm{SFT}})/2$.
Together, these signals characterize whether the base model lacks relevant knowledge, holds conflicting priors, or undergoes insufficient or unstable updates during fine-tuning. In the next subsection, we use these diagnostics to analyze unlearned samples and map them to distinct sources of incomplete learning.

\subsection{Unlearned Sample Processing}
\label{subsec:framework}

\begin{figure}
\centering
\includegraphics[width=0.48\textwidth]{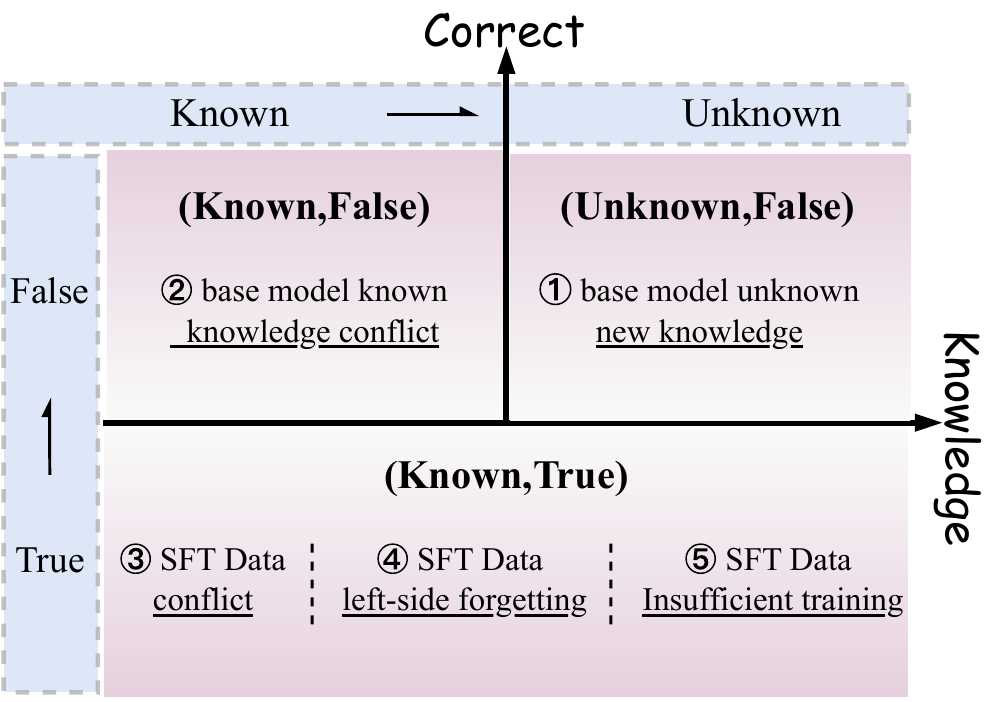}
\caption{\label{fig:example2}Unlearned sample attribution framework.}
\end{figure}

To systematically analyze the Incomplete Learning Phenomenon (ILP), we introduce a unified pipeline that operates at the level of individual supervised instances.
The core objective is not merely to improve aggregate performance, but to determine \emph{why} specific SFT samples remain unlearned after convergence.
Figure~\ref{fig:example2} illustrates the overall attribution process.
The pipeline begins by identifying unlearned samples via post-SFT evaluation.
Each such sample is then sequentially examined under a set of diagnostic tests, each corresponding to a hypothesized source of incomplete learning.
Importantly, the mitigation strategies described below are not positioned as general-purpose solutions; instead, they serve as controlled interventions to validate the causal relevance of each attributed factor.

\subsubsection{Base Model Knowledge Limitations}

The first step of our framework focuses on identifying knowledge blind spots in the base model. We begin by detecting unlearned samples from the SFT dataset and extracting their underlying factual content using OpenIE tools.\footnote{https://nlp.stanford.edu/software/openie.html} Each sample is converted into a set of subject--predicate--object triplets, forming a candidate knowledge set $\mathcal{K}_\text{cand} = \{(h, r, t)\}$.

To quantify whether a knowledge triplet is sufficiently covered by the base model, we adopt BoN sampling and the $\text{pass@N}$ metric as probing mechanisms. Intuitively, if the model repeatedly fails to produce correct answers even under multiple sampling attempts, the corresponding knowledge is likely missing rather than poorly optimized. Formally, we define the set of blind knowledge as:
\begin{equation}
\begin{aligned}
\mathcal{K}_\text{blind} = \{k \mid \text{pass@10}(k) < 0.2 \\ \land \text{BoN-5 Acc}(k) < 0.1\}.
\end{aligned}
\end{equation}
This criterion filters out cases where errors are attributable to stochasticity or reasoning noise, retaining only those samples that reflect systematic knowledge gaps.
Once blind knowledge is identified, we expand the corresponding background information by querying multiple external sources, including WikiData APIs, Google Search, and the OpenAI-o1 API. For each unknown entity, we retrieve an average of $20 \pm 1.1$ related documents, covering definitions, relations, and contextual usage. This multi-source aggregation mitigates bias from any single knowledge provider and improves factual completeness.
The resulting knowledge-augmented corpus $\mathcal{C}_\text{aug}$ is then mixed with a general-domain corpus to perform continued pre-training. The mixed corpus as:
\begin{equation}
\mathcal{C}_\text{mix} = 0.8\mathcal{C}_\text{general} + 0.2\mathcal{C}_\text{aug},
\end{equation}
where $\mathcal{C}_\text{general}$ consists of standard pre-training data such as OpenWebText and BookCorpus. This design explicitly balances knowledge injection and distributional stability, enabling the model to acquire missing facts without degrading its general language understanding capabilities.
After CPT, we reapply SFT using the original SFT dataset and evaluate the updated model. Improvements are measured using accuracy and $\text{pass@N}$ metrics, allowing us to isolate gains attributable to knowledge completion rather than optimization artifacts. As shown in Figure~\ref{fig:result2}, this procedure consistently improves downstream performance across medical, legal, and financial benchmarks, validating that incomplete learning in SFT can often be traced back to knowledge deficiencies in the base model.

\subsubsection{Conflicts Between SFT and Base Model}

Beyond missing knowledge, we observe another failure mode in which the base model exhibits strong but incorrect beliefs that conflict with SFT supervision. Such conflicts are particularly problematic because high-confidence errors tend to resist correction during fine-tuning, leading to unstable or slow convergence.
To systematically identify these cases, we prompt the base model to answer multiple-choice questions from the SFT dataset and extract the probability of the predicted answer token. Let $P_{\text{model}}(y \mid x)$ denote the model’s confidence for input $x$ and predicted answer $y$. A sample is flagged as a high-confidence error if the model strongly prefers an incorrect answer:
\begin{equation}
\begin{aligned}
\text{Error}(x, y) = 
\end{aligned}
\begin{cases}
\begin{aligned}
1, & P(y|x) > T  \text{ and } y \neq y_{\text{SFT}}, \\
0, & \text{otherwise}.
\end{aligned}
\end{cases}
\end{equation}
Here, $T$ denotes a predefined confidence threshold, and $y_{\text{SFT}}$ is the ground-truth label provided by the SFT data. Samples satisfying this condition form a high-confidence error set $\mathcal{E}$, representing explicit knowledge conflicts between the base model and supervision.
To resolve these conflicts, we follow the same knowledge augmentation and CPT procedure described above. Specifically, authoritative external sources such as Wikipedia and domain-specific corpora are used to retrieve verified information corresponding to conflicting samples. Continued pre-training on this curated corpus realigns the model’s internal knowledge representations, reducing resistance to subsequent SFT updates.





\subsubsection{Knowledge Conflicts Within SFT Data}

Incomplete learning may also originate from inconsistencies internal to the SFT dataset itself.
When semantically similar inputs are associated with contradictory labels, the model receives an incoherent learning signal, limiting convergence on affected samples.
We detect such conflicts by computing semantic similarity between sample pairs.
If $\text{Sim}(s_i, s_j) > X$, the pair is treated as potentially conflicting.
To determine correctness, we employ GPT\cite{openai2023gpt4},deepseek\cite{deepseekai2025deepseekv3technicalreport} as an external evaluator.
If one sample is judged incorrect, it is removed; if both are judged correct, the pair is retained but treated as incompatible during training.
Rather than discarding valid supervision, we assign conflicting samples to separate training buckets, ensuring they do not co-occur within the same mini-batch.
This bucket assignment is periodically re-evaluated every $K$ training steps to reflect the model’s evolving competence.
Observed reductions in error rates on these samples after bucketing indicate that internal data conflict, rather than representational insufficiency, was the primary cause of incomplete learning.



\begin{figure*}
\centering
\includegraphics[width=1.0\textwidth]{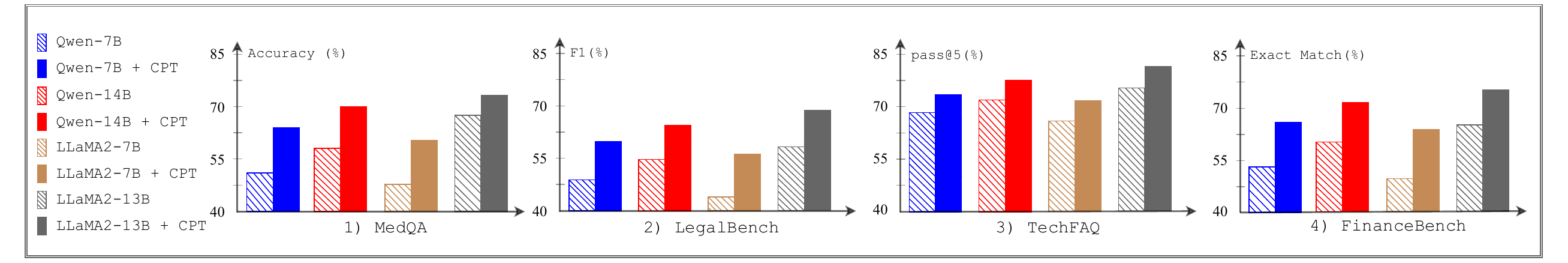}
\caption{\label{fig:result2}Performance improvements achieved by introducing Continued Pre-Training (CPT). Results demonstrate consistent accuracy gains across the medical (MedQA), legal (LegalBench), and financial (FinanceBench) domains.
}
\end{figure*}

\subsubsection{Left-side Forgetting}

Another manifestation of incomplete learning appears as left-side forgetting. When SFT datasets are concatenated or processed sequentially, we observe a systematic bias toward recently seen data.
By reversing dataset order and tracking per-dataset accuracy, we find that earlier samples are progressively overshadowed, consistent with left-side forgetting \cite{li2024examining}.
To mitigate this effect, we apply random shuffling across the entire SFT dataset and introduce a dynamic re-sampling mechanism.
At regular intervals of $K$ steps, validation accuracy is monitored for each data subset.
If a significant drop is detected, samples from the affected subset are temporarily upweighted.
This strategy, detailed in Algorithm~\ref{alg:dynamic_resampling}, serves to test whether incomplete learning arises from training order effects rather than intrinsic difficulty.

\subsubsection{Insufficient Training}

Finally, incomplete learning can arise from insufficient optimization, where a fixed number of training epochs fails to accommodate datasets of varying complexity. To address this, we adopt a progressive epoch increment strategy inspired by early stopping \cite{prechelt2002early}. Training begins with a minimal epoch count $E_{\min}$ and incrementally increases until validation performance ceases to improve. The stopping condition is defined as
\begin{equation}
\mathcal{C}\text{stop} = \mathbb{I}\left(\mathcal{L}\text{val}^{(e)} > \mathcal{L}_\text{val}^{(e-1)} + \delta\right),
\label{eq:stop-condition}
\end{equation}
where $\delta = 0.01$ prevents premature termination due to noise. This adaptive strategy ensures sufficient learning while avoiding overfitting, and its implementation is detailed in Algorithm~\ref{alg:progressive_epoch}.

\begin{table*}[t]
\centering
\small
\renewcommand{\arraystretch}{0.9}
\setlength{\tabcolsep}{4pt}

\begin{tabularx}{\textwidth}{l *{4}{>{\centering\arraybackslash}X}}
\toprule
\textbf{Model} &
\textbf{ARC} &
\textbf{CommonQA} &
\textbf{SocialIQA} &
\textbf{MedMCQA} \\
\midrule
Qwen 7B
& 68.1 $\rightarrow$ 70.9 \improve{2.8}
& 74.5 $\rightarrow$ 76.9 \improve{2.4}
& 70.3 $\rightarrow$ 72.2 \improve{1.9}
& 61.2 $\rightarrow$ 63.7 \improve{2.5} \\

Qwen 14B
& 71.2 $\rightarrow$ 73.7 \improve{2.5}
& 76.8 $\rightarrow$ 79.1 \improve{2.3}
& 72.1 $\rightarrow$ 74.0 \improve{1.9}
& 63.1 $\rightarrow$ 65.3 \improve{2.2} \\

LLaMA2 7B
& 66.3 $\rightarrow$ 68.8 \improve{2.5}
& 72.8 $\rightarrow$ 75.3 \improve{2.5}
& 68.9 $\rightarrow$ 71.1 \improve{2.2}
& 58.9 $\rightarrow$ 61.4 \improve{2.5} \\

LLaMA2 13B
& 69.1 $\rightarrow$ 71.3 \improve{2.2}
& 75.2 $\rightarrow$ 77.3 \improve{2.1}
& 70.5 $\rightarrow$ 72.1 \improve{1.6}
& 61.1 $\rightarrow$ 63.0 \improve{1.9} \\
\bottomrule
\end{tabularx}

\caption{
Accuracy (\%) before and after Continued Pre-Training (CPT) on four knowledge-intensive benchmarks.
Improvements brought by CPT are highlighted in color and remain consistent across model sizes and domains.
}
\label{tab:cpt_accuracy}
\end{table*}

\begin{table*}[t]
\centering
\small
\renewcommand{\arraystretch}{0.9}
\setlength{\tabcolsep}{4pt}

\begin{tabularx}{\textwidth}{l *{4}{>{\centering\arraybackslash}X}}
\toprule
\textbf{Experiment} &
\textbf{Qwen-7B} &
\textbf{Qwen-14B} &
\textbf{LLaMA-7B} &
\textbf{LLaMA-13B} \\
\midrule
Knowledge conflict
& 82.3 $\rightarrow$ 85.1 \improve{2.8}
& 84.5 $\rightarrow$ 87.2 \improve{2.7}
& 81.8 $\rightarrow$ 84.3 \improve{2.5}
& 83.6 $\rightarrow$ 86.5 \improve{2.9} \\

Left-side forgetting
& 78.5 $\rightarrow$ 79.8 \improve{1.3}
& 79.3 $\rightarrow$ 80.5 \improve{1.2}
& 77.9 $\rightarrow$ 79.1 \improve{1.2}
& 78.7 $\rightarrow$ 80.2 \improve{1.5} \\

Insufficient learning
& 88.2 $\rightarrow$ 90.1 \improve{1.9}
& 89.5 $\rightarrow$ 91.3 \improve{1.8}
& 87.8 $\rightarrow$ 89.7 \improve{1.9}
& 88.9 $\rightarrow$ 90.8 \improve{1.9} \\
\bottomrule
\end{tabularx}

\caption{
Performance improvements across baseline models after applying optimization strategies for resolving knowledge conflicts, mitigating left-side forgetting, and addressing insufficient learning.
}
\label{tab:optimization_results2}
\end{table*}

\section{Results Analysis}

\subsection{Base Model Knowledge Enhancement}
To address pre-training knowledge gaps, we employ Continued Pre-Training (CPT), described in Appendix~\ref{cpt}. As illustrated in Figure~\ref{fig:result2}, CPT consistently improves accuracy on domain-specific benchmarks, including MedQA \cite{jin2020disease}, LegalBench \cite{guha2023legalbench,koreeda2021contractnli}, and FinanceBench \cite{islam2023financebench} for models such as Qwen \cite{qwen} and LLaMA2 \cite{llama2}. Accuracy gains range from 9.4\% to 14.1\% (e.g., +12.5\% on MedQA), demonstrating CPT's effectiveness in filling critical knowledge gaps that standard SFT fails to capture. Notably, simply extending SFT epochs leads to only marginal improvements, underscoring that missing foundational knowledge cannot be addressed through prolonged fine-tuning alone.
Further validation with OLMo2-7B \cite{olmo20242olmo2furious} (Appendix~\ref{olmo}) shows similar trends: CPT significantly enhances performance in domains where the base model initially exhibited high 'Knowledge Non-Existence Rates'. While targeted knowledge injection sometimes interacts with generalization, careful corpus balancing mitigates negative effects, indicating that CPT can selectively improve domain-specific knowledge without undermining overall language understanding. Collectively, these results highlight CPT as a necessary step for bridging knowledge deficits in LLMs prior to SFT.

\subsection{Knowledge Conflict Calibration}
High-confidence conflicts between pre-trained knowledge and SFT supervision pose another obstacle to complete learning. To resolve this, we apply a CPT-based calibration strategy (Appendix~\ref{conflicts}). Table~\ref{tab:cpt_accuracy} demonstrates consistent accuracy improvements across models(Qwen-7B/14B, LLaMA2-7B, and LLaMA2-13B) on diverse benchmarks. Gains range from +1.6\% (LLaMA2-13B on SocialIQA) to +2.8\% (Qwen-7B on ARC), with additional improvements for other datasets (e.g., +2.5\% for Qwen-14B on ARC, +2.1\% for LLaMA2-13B on CommonQA). These improvements correspond to a marked reduction in high-confidence SFT conflicts, confirming that targeted CPT effectively aligns the model's predictions with supervised knowledge.
Case studies on OLMo2-7B reveal that CPT recalibrates conflict points where pre-trained knowledge previously overrode SFT supervision. This demonstrates CPT's dual role: both filling missing knowledge and mitigating entrenched misbeliefs. Overall, CPT provides a systematic mechanism to harmonize pre-training and supervised signals, which is essential for reducing incomplete learning arising from knowledge conflicts.

\subsection{SFT Knowledge Conflict Resolution}

Internal conflicts within the SFT dataset can also induce incomplete learning, particularly when semantically similar or nearly identical inputs are paired with contradictory labels—introducing noise that confuses the optimization process. To mitigate this, we propose a two-stage approach based on conflict detection followed by dynamic bucketing, which groups potentially conflicting examples into separate training batches while preserving all valid supervision signals (see Appendix~\ref{sft_conflict} for implementation details). As demonstrated in Table~\ref{tab:optimization_results2}, this strategy yields consistent and substantial performance gains on mixed-domain SFT datasets: for instance, Qwen-7B improves from 82.3\% to 85.1\% (+2.8\%) and Qwen-14B from 84.5\% to 87.2\% (+2.7\%). Comparable improvements are observed across LLaMA model variants, confirming the generality of the method. By isolating conflicting samples into distinct batches rather than discarding them, the model retains access to valuable supervisory information and learns more robust representations from complex, real-world SFT data. Ablation studies reported in Table~\ref{tab:optimization_results3} further validate that dynamic bucketing significantly outperforms naive conflict resolution strategies—such as removing all samples flagged as conflicting—which often eliminate informative examples and inadvertently reduce the overall learning capacity of the model.

\subsection{Alleviating Left-Side Forgetting}

Left-side forgetting, where early-learned SFT knowledge is progressively overshadowed or even overwritten during sequential training on multi-task or mixed-domain data, represents another critical source of incomplete learning. To counteract this temporal bias, we employ a joint strategy of global shuffling—randomizing the entire training sequence across epochs—together with dynamic resampling that adaptively upweights earlier examples throughout training (Appendix~\ref{left_side_forgetting}). This dual approach ensures that initial knowledge remains actively reinforced rather than diluted by later batches. As shown in Table~\ref{tab:optimization_results2}, this leads to consistent accuracy improvements on mixed datasets: Qwen-7B rises from 78.5\% $\rightarrow$ 79.8\%, and Qwen-14B from 79.3\% $\rightarrow$ 80.5\%. More importantly, ROUGE-L scores on the first 10\% of summarization data—the segment most vulnerable to left-side forgetting—increased significantly by +29\% (from 0.41 $\rightarrow$ 0.53, Table~\ref{tab:rouge_results4}), demonstrating robust preservation of early-acquired capabilities. These results confirm that the combination of dynamic resampling and global shuffling effectively mitigates progressive knowledge decay while minimally interfering with the acquisition of later-stage tasks.

\subsection{Alleviating Insufficient Learning}
Insufficient optimization, particularly for rare, long-tail, or structurally complex patterns in SFT datasets, is a key contributor to incomplete learning. Standard fixed-epoch training often terminates before such difficult examples receive adequate signal, leaving residual errors that degrade model reliability. To address this limitation, we employ a Progressive Epoch Increment strategy combined with validation-driven early stopping (Appendix~\ref{insufficient}), which dynamically adapts training duration per dataset based on real-time validation performance. This adaptive schedule allocates additional epochs only when marginal gains are observed, ensuring that underrepresented or challenging examples receive sufficient gradient updates while simultaneously preventing overfitting through timely termination. As shown in Table~\ref{tab:optimization_results2}, Qwen-7B accuracy on underlearned tasks increases from 88.2\% to 90.1\% (+1.9\%), with comparable gains observed across Qwen-14B, LLaMA-7B, and LLaMA-13B, whose improvements range from +1.0\% to +1.9\%. These results demonstrate that adaptive training duration effectively closes persistent learning gaps for difficult data patterns, thereby enhancing both model completeness and robustness—without compromising generalization on broader benchmarks or incurring unnecessary computational cost.
\section{Conclusion}

In this paper, we systematically investigate the “Incomplete Learning Phenomenon” (ILP) in supervised fine-tuning (SFT) of large language models (LLMs) and identify five major contributing factors: (1) limitations in pre-training knowledge that hinder downstream adaptation, (2) conflicts between SFT data and the base model’s priors, (3) internal inconsistencies within the SFT dataset itself, (4) left-side forgetting during sequential training, and (5) insufficient optimization due to inadequate training duration or data exposure.
To address these interrelated challenges, we introduce a unified mitigation framework integrating pre-training enhancement, conflict-aware data processing, dynamic bucketing, data resampling, and adaptive epoch augmentation. Extensive experiments with multiple LLMs across diverse datasets demonstrate that these strategies collectively and effectively mitigate ILP, resulting in significant improvements not only in the model’s mastery of SFT-specific knowledge but also in generalization performance on standard evaluation benchmarks.

\section{Limitations}
Despite the effectiveness and breadth of our proposed framework for addressing the Incomplete Learning Phenomenon in Supervised Fine-Tuning (SFT), several limitations warrant further investigation:

\noindent \textbf{Complexity of Conflict Detection:} While we have proposed strategies for detecting and resolving knowledge conflicts (both between pre-training and SFT data, and within the SFT data itself), the current approach depends on high-quality annotations and reliable external tools (e.g., for domain verification). Inconsistent or noisy data sources may reduce conflict detection accuracy, leading to potentially suboptimal or partial conflict resolution.

\noindent \textbf{Dependency on Quality Pre-training Data:} Our method presupposes that injecting additional knowledge or updates into the pre-training phase will robustly bridge knowledge gaps. However, if the supplementary corpus is itself noisy or a biased representative, those newly introduced biases or errors could propagate through subsequent fine-tuning stages, diminishing overall performance gains.

\noindent \textbf{Computational Overheads:} The inclusion of pre-training enhancement and knowledge resampling increases training time and resource consumption.  Particularly at the multi-billion parameter level LLMs, amplify computational demands, raising concerns about feasibility for organizations with limited hardware or training budgets.

Overall, while our proposed framework alleviates many inherent challenges of SFT in large language models, the actual fine-grained calibration remains underexplored.

\bibliography{custom}

\clearpage
\appendix

\newpage

\section{Base Model Knowledge Limitations Supplement}
\label{cpt}

\begin{table}[ht]
\centering
\renewcommand{\arraystretch}{1.0} 
\setlength{\tabcolsep}{3pt} 
\scalebox{0.85}{
\begin{tabular}{l l c c}
\toprule
\textbf{Dataset} & \textbf{Domain} & \textbf{Sample Size} & \textbf{Evaluation} \\
\midrule
MedQA & Medical & 12,873 & Accuracy \\
LegalBench & Legal & 8,452 & F1-score \\
TechFAQ & Technology & 5,621 & pass@5 \\
FinanceBench & Finance & 10,150 & EM \\
\bottomrule
\end{tabular}
}
\caption{\label{s1}Statistics of the SFT datasets and their corresponding evaluation metrics.}
\end{table}

\subsection{Experimental Datasets}
As shown in Table \ref{s1}, the experiment employs standard test sets from diverse domains to validate the consistency and efficacy of the method in augmenting knowledge across various fields. The specific datasets utilized are as follows:

\noindent \textbf{MedQA:}\cite{jin2020disease} comprises question-and-answer pairs in the medical domain, assessing the model's proficiency in medical expertise.

\noindent \textbf{LegalBench:}\cite{guha2023legalbench} Focused on legal knowledge and question-answering, this dataset evaluates the model's capacity to comprehend and interpret legal statutes and case law.

\noindent \textbf{TechFAQ:}\cite{liang2021rdrop} encompasses common issues in the information technology sector, testing the model's grasp of technical knowledge, such as programming and network security.

\noindent \textbf{FinanceBench
:}\cite{islam2023financebench} Centered on financial topics, which measures the model's understanding of economics and financial accounting.

\subsection{Experimental Baselines}
To evaluate the impact of different models before and after addressing the knowledge gap, employs the following representative LLMs as baselines: \textbf{qwen-7b}, \textbf{qwen-14b}, \textbf{llama2-8B}, \textbf{llama2-13B}.
These baselines vary in parameter scales, allowing for a more comprehensive assessment of the adaptability and enhancement effects of the proposed method across each model.

\subsection{Algorithm of Knowledge-Enhanced Continue Pre-training}
Our "knowledge-enhanced continual pre-training" method, illustrated in Algorithm \ref{alg:knowledge_enhanced_pretrain}, addresses pre-training knowledge limitations.
The process starts by identifying SFT samples the base model fails to learn. These are processed via OpenIE\footnote{\url{https://nlp.stanford.edu/software/openie.html}}) into candidate knowledge triplets ($\mathcal{K}_\text{cand}$). As outlined in Step 1 of Algorithm \ref{alg:knowledge_enhanced_pretrain}, model proficiency on $\mathcal{K}_\text{cand}$ is assessed using pass@N and BoN accuracy to identify 'blind knowledge triplets' ($\mathcal{K}_\text{blind}$).

\begin{algorithm}[t]
\caption{Knowledge Continue Pre-train}
\label{alg:knowledge_enhanced_pretrain}
\begin{algorithmic}[1]

\State \textbf{Require:} SFT dataset $\mathcal{D}_{\text{SFT}}$, base model $M_{\text{base}}$
\State \textbf{Ensure:} Optimized base model

\vspace{0.3em}
\State \textbf{Step 1: Identify Knowledge Gaps}
\State Extract unlearned samples into knowledge graph triples $\mathcal{K}_\text{cand} = \{(h,r,t)\}$.
\State Use BoN and $\text{pass@N}$ indicators to locate blind areas:
\begin{align*}
\mathcal{K}_\text{blind} = \{k \mid & \text{pass@10}(k) < 0.2 \\
& \land \text{BoN-5 Acc}(k) < 0.1\}.
\end{align*}

\State \textbf{Step 2: Collect External Knowledge}
\For{Each blind area entity $e \in \mathcal{K}_\text{blind}$}
    \State Use WikiData, Google Search, and other extended background knowledge to build corpus $\mathcal{C}_\text{aug}$.
\EndFor

\State \textbf{Step 3: Continue Pre-training}
\State Mix general data with augmented corpus:
\begin{equation*}
\mathcal{C}_\text{mix} = 0.8\mathcal{C}_\text{general} + 0.2\mathcal{C}_\text{aug}.
\end{equation*}
\State Continue pre-training with $\mathcal{C}_\text{mix}$.

\State \textbf{Step 4: Validate with SFT}
\State Perform SFT on the updated model and evaluate the performance improvement.

\State \Return Optimized base model

\end{algorithmic}
\end{algorithm}

Step 2 details the construction of an augmented corpus ($\mathcal{C}_\text{aug}$) for these deficient areas using external resources (WikiData API, Google Search, OpenAI API). We prioritize collecting foundational information and conceptual explanations pertinent to the knowledge area, deliberately avoiding direct content from the original unlearned SFT samples to ensure CPT supplements understanding. This yields approximately $20 \pm 1.1$ documents per area.

In Step 3, $\mathcal{C}_\text{aug}$ is mixed with a general pre-training dataset $\mathcal{C}_{general}$ (e.g., $0.8\mathcal{C}_\text{general} + 0.2\mathcal{C}_\text{aug}$, found effective in experiments – discussion in Appendix V), and the base model undergoes CPT on this mix. Step 4 validates the enhanced model via SFT and subsequent evaluation on standard benchmarks, demonstrating improved knowledge coverage and performance.

\begin{table*}[h!]
\centering
\small
\renewcommand{\arraystretch}{1.2}
\setlength{\tabcolsep}{6pt}

\begin{tabularx}{\textwidth}{l l *{3}{>{\centering\arraybackslash}X}}
\toprule
\textbf{Dataset} & \textbf{Method} & \textbf{Acc (\%)} & \textbf{pass@10} & \textbf{Learn Rate (\%)} \\
\midrule
MedQA
& Baseline (2 epoch SFT)
& 0.0 & 0.0 & 65.3 \\
& \quad Increase epoch to 10
& \textcolor{red}{+1.2} & \textcolor{red}{+1.5} & 66.8 \\
& \quad Continued pre-training + SFT
& \textcolor{red}{+8.3} & \textcolor{red}{+10.5} & 82.1 \\
\midrule
LegalBench
& Baseline (2 epoch SFT)
& 0.0 & 0.0 & 62.5 \\
& \quad Increase epoch to 10
& \textcolor{red}{+1.0} & \textcolor{red}{+1.3} & 63.8 \\
& \quad Continued pre-training + SFT
& \textcolor{red}{+7.9} & \textcolor{red}{+9.8} & 80.5 \\
\midrule
TechFAQ
& Baseline (2 epoch SFT)
& 0.0 & 0.0 & 68.1 \\
& \quad Increase epoch to 10
& \textcolor{red}{+1.4} & \textcolor{red}{+1.8} & 69.5 \\
& \quad Continued pre-training + SFT
& \textcolor{red}{+8.7} & \textcolor{red}{+11.2} & 83.6 \\
\midrule
FinanceBench
& Baseline (2 epoch SFT)
& 0.0 & 0.0 & 63.8 \\
& \quad Increase epoch to 10
& \textcolor{red}{+1.1} & \textcolor{red}{+1.6} & 65.4 \\
& \quad Continued pre-training + SFT
& \textcolor{red}{+8.0} & \textcolor{red}{+10.1} & 81.7 \\
\bottomrule
\end{tabularx}

\caption{
Comparison of SFT performance by increasing training epochs and applying continued pre-training (CPT + SFT) across four datasets.
Performance improvements are highlighted in red.
}
\label{tab:experiment_results}
\end{table*}

\subsection{Verification of the Unlearnability of Knowledge Blind Spots by Increasing SFT Epochs}

\subsubsection{Experimental Design}

To further investigate the characteristics of knowledge blind spots in the base model, we designed a comparative experiment, addressing the following questions: \underline{Can the knowledge gaps of the} \underline{base model be filled by increasing the number of} \underline{ training rounds (epochs) of SFT? }
The experimental results are presented in Table \ref{tab:experiment_results}. 

The results indicate that increasing the number of epochs in Supervised Fine-Tuning shows limited effectiveness in improving the model's performance in areas where it lacks knowledge. For example, in the MedQA dataset, extending the training from 2 to 10 epochs only marginally increased the coverage rate of knowledge blind spots from 65.3\% to 66.8\%. This suggests that simply increasing the number of SFT training epochs does not significantly address the knowledge gaps in the base model.
\begin{table}[ht]
\centering
\renewcommand{\arraystretch}{1.0} 
\setlength{\tabcolsep}{4pt} 
\scalebox{0.9}{
\begin{tabular}{l l c c}
\toprule
\textbf{Dataset} & \textbf{Field} & \textbf{Size} & \textbf{Evaluation} \\
\midrule
ARC & Science & 7,787 & Accuracy \\
Common & Commonsense & 12,247 & Accuracy \\
SocialIQA & Social & 33,410 & Accuracy \\
MedMCQA & Medical & 187,995 & Accuracy \\
\bottomrule
\end{tabular}
}
\caption{\label{l1}SFT Dataset statistics and evaluation index.}
\end{table}
In contrast, continuing pre-training with knowledge enhancement significantly improves the model's ability to cover these blind spots. For instance, in the TechFAQ dataset, the coverage rate increased from 68.1\% to 83.6\%. This underscores the importance of incorporating external knowledge during the pre-training stage to enable the base model to acquire missing knowledge, which is critical for effective SFT.
Furthermore, the experimental results reveal a certain "unlearnability" of the base model's knowledge blind spots. Even with additional SFT training epochs, the model struggles to master the missing knowledge. This highlights the importance of addressing these gaps during the pre-training stage.

This finding emphasizes the critical role of knowledge injection in the pre-training stage in the optimization process of large-scale language models. For the knowledge blind spot of the base model, it is not enough to rely solely on SFT. External knowledge must be introduced through a continued pre-training phase for optimizing large-scale language models.

\section{Conflicts Between SFT and Base Model Supplement}
\label{conflicts}

\begin{table*}[h!]
\centering
\small
\renewcommand{\arraystretch}{1.0}
\setlength{\tabcolsep}{4pt}

\begin{tabularx}{\textwidth}{l *{4}{>{\centering\arraybackslash}X}}
\toprule
\textbf{Model} &
ARC (+CPT) &
CommonQA (+CPT) &
SocialIQA (+CPT) &
MedMCQA (+CPT) \\
\midrule
Qwen 7B
& 12.3\% $\rightarrow$ 8.8\% \improved{-3.5}
& 10.1\% $\rightarrow$ 7.3\% \improved{-2.8}
& 11.7\% $\rightarrow$ 8.8\% \improved{-2.9}
& 14.5\% $\rightarrow$ 10.7\% \improved{-3.8} \\

Qwen 14B
& 11.2\% $\rightarrow$ 8.0\% \improved{-3.2}
& 9.3\% $\rightarrow$ 6.8\% \improved{-2.5}
& 10.8\% $\rightarrow$ 8.2\% \improved{-2.6}
& 13.1\% $\rightarrow$ 9.6\% \improved{-3.5} \\

LLaMA2 7B
& 13.1\% $\rightarrow$ 9.5\% \improved{-3.6}
& 10.7\% $\rightarrow$ 7.7\% \improved{-3.0}
& 12.3\% $\rightarrow$ 9.2\% \improved{-3.1}
& 15.2\% $\rightarrow$ 11.2\% \improved{-4.0} \\

LLaMA2 13B
& 12.5\% $\rightarrow$ 9.2\% \improved{-3.3}
& 9.8\% $\rightarrow$ 7.1\% \improved{-2.7}
& 11.5\% $\rightarrow$ 8.7\% \improved{-2.8}
& 14.3\% $\rightarrow$ 10.6\% \improved{-3.7} \\
\bottomrule
\end{tabularx}

\caption{
Relative reduction of conflict rates on the SFT dataset before and after Continued Pre-Training (CPT) for each model.
Negative improvements indicate a decrease in conflict rates, consistently observed across benchmarks and model sizes.
}
\label{tab:r2}
\end{table*}

\subsection{Experimental Datasets}
As presented in Table \ref{l1}, the experiment utilizes datasets from multiple domains to validate the consistency and effectiveness of the method augmenting knowledge across various fields, as: 

\noindent \textbf{ARC(AI2 Reasoning Challenge):}\cite{allenai:arc} This dataset comprises science-related questions, categorized into easy and challenging levels, focusing on the model's reasoning capabilities and knowledge in the scientific domain.

\noindent \textbf{CommonsenseQA:}\cite{talmor-etal-2019-commonsenseqa} A multiple-choice dataset designed for commonsense reasoning, requiring the model to possess extensive commonsense knowledge. It evaluates the model's performance in handling questions that demand background knowledge and logical reasoning.

\noindent \textbf{SocialIQA:}\footnote{https://huggingface.co/datasets/allenai/social\_i\_qa} This dataset covers questions related to social commonsense reasoning, involving emotions, social norms, and interpersonal interactions. It focuses on the model's understanding of social contexts and human behavior.

\noindent \textbf{MedMCQA:}\cite{pmlr-v174-pal22a} A multiple-choice dataset in the medical field, encompassing a wide range of medical knowledge and clinical reasoning. It tests the model's ability to handle complex medical questions and support clinical decision-making.

\subsection{Algorithm for Resolving Calibration Conflicts Between SFT and Base Model }

The optimization strategy, which leverages high-confidence error detection and CPT, is outlined in Algorithm \ref{alg:knowledge_alignment}. Initially, high-confidence error samples are identified by comparing the model's predictions with the ground truth labels in the SFT dataset. If the model’s confidence in an incorrect prediction surpasses a predefined threshold, the sample is classified as a high confidence error and included to the error set $\mathcal{E}$.
Subsequently, for each identified error sample, relevant knowledge is gathered from external sources, such as WikiData or other knowledge repositories, to construct an enhanced knowledge corpus $\mathcal{K}_i$. This step ensures that the model acquires additional context and information to rectify its errors. Furthermore, domain-specific databases and academic papers are considered to be utilized for a more comprehensive knowledge base. 
Finally, the enhanced knowledge corpus is integrated with the general pre-training dataset at a specified ratio $\alpha$, and the model undergoes continued pre-training with this combined dataset. This approach aims to enhance the model's accuracy by targeting specific areas where it previously made high-confidence errors. The model's effectiveness is subsequently evaluated through  SFT and validation steps to confirm performance improvements.

The experimental results are presented in Table \ref{tab:r2}. From the perspective of the reduction in the data conflict rate, all models across the four datasets exhibit a significant decrease in conflict rates. This indicates that CPT effectively mitigates the knowledge conflicts between the model and the SFT data. Notably, Qwen 14B outperforms other models in reducing the conflict rate, likely due to its larger parameter scale. Furthermore,  the most substantial reduction in conflict rate is observed on the MedMCQA dataset, suggesting that external knowledge retrieval and continued pre-training have a particularly pronounced effect on knowledge calibration in the medical domain.
In summary, the observed reduction in the data conflict rate further validates the effectiveness of the high-confidence error detection-based method. CPT significantly mitigates model knowledge conflicts, thereby enhancing the model's performance and reliability.

\begin{algorithm}[t]
\caption{Optimization Strategy Based on High-Confidence Error Detection and Continued Pre-training}
\label{alg:knowledge_alignment}
\begin{algorithmic}[1]

\State \textbf{Input:} 
SFT dataset $\mathcal{D}_{\text{SFT}}$; 
base model $M_{\text{base}}$; 
confidence threshold $T_{\text{conf}}$; 
external knowledge source $\mathcal{K}$

\State \textbf{Output:} Optimized base model

\vspace{0.5em}
\State Initialize high-confidence error set $\mathcal{E} \leftarrow \emptyset$

\ForAll{$(x, y_{\text{SFT}}) \in \mathcal{D}_{\text{SFT}}$}
    \State Obtain model prediction distribution $P_{\text{model}}(y \mid x)$ using $M_{\text{base}}$
    \If{$P_{\text{model}}(y \mid x) > T_{\text{conf}}$ \textbf{and} $y \neq y_{\text{SFT}}$}
        \State $\mathcal{E} \leftarrow \mathcal{E} \cup \{(x, y_{\text{SFT}})\}$
    \EndIf
\EndFor

\ForAll{$e_i \in \mathcal{E}$}
    \State Retrieve relevant knowledge from $\mathcal{K}$ and construct a knowledge-enhanced corpus $K_i$
\EndFor

\State Mix the aggregated knowledge-enhanced corpus with the general pre-training data according to the ratio $\alpha$
\State Continue pre-training $M_{\text{base}}$ on the mixed corpus

\State \Return the optimized base model

\end{algorithmic}
\end{algorithm}

\section{Knowledge Conflicts Between SFT Data Supplement}
\label{sft_conflict}
\subsection{Experimental Datasets}
\begin{table}
\centering
\renewcommand{\arraystretch}{1.0} 
\setlength{\tabcolsep}{1pt} 
\scalebox{0.87}{
\begin{tabular}{l l c c}
\toprule
\textbf{Dataset} & \textbf{Field} & \textbf{Size} & \textbf{Evaluation} \\
\midrule
SQuAD & General & 150,000 &  EM \\
CoQA & Conversational & 127,000 & F1 \\
TriviaQA & Trivia & 95,000 & Acc \\
Natural Questions & Web Search & 307,373 & EM \\
\bottomrule
\end{tabular}
}
\caption{\label{l3}SFT Dataset statistics and evaluation index.}
\end{table}

To evaluate the model's performance in knowledge conflict scenarios, we utilize a diverse set of question-answering datasets, each designed to test different aspects of the model's knowledge and reasoning capabilities, and summarized in Table \ref{l3}:

\noindent \textbf{SQuAD (Stanford QA Dataset):}\cite{rajpurkar-etal-2016-squad} A widely-used dataset for reading comprehension, focusing on extracting answers from provided passages. This dataset evaluates the model's ability to handle context-specific information and resolve potential conflicts within the text.

\noindent \textbf{CoQA
 (Conversational Question Answering Dataset):}\cite{reddy-etal-2019-coqa} A dataset designed for conversational question answering, requiring the model to maintain context across multiple dialogue turns. This tests the model's ability to ensure knowledge consistency in dynamic interactions.

\noindent \textbf{TriviaQA\cite{2017arXivtriviaqa}:} A large-scale dataset containing trivia questions spanning a wide range of topics. It challenges the model's general knowledge and its ability to resolve conflicts between different information sources.

\noindent \textbf{Natural Questions(NQ):}\cite{47761} based on user queries from Google Search, focusing on open-domain question answering. This dataset evaluates the model's capacity to integrate information from diverse knowledge sources.

\begin{table*}[h!]
\centering
\small
\renewcommand{\arraystretch}{1.25}
\setlength{\tabcolsep}{4pt}

\begin{tabularx}{\textwidth}{l *{4}{>{\centering\arraybackslash}X}}
\toprule
\textbf{Experiment} &
\textbf{Qwen-7B} &
\textbf{Qwen-14B} &
\textbf{LLaMA-7B} &
\textbf{LLaMA-13B} \\
\midrule
Knowledge conflict
& 82.3\% $\rightarrow$ 85.1\% \improve{2.8}
& 84.5\% $\rightarrow$ 87.2\% \improve{2.7}
& 81.8\% $\rightarrow$ 84.3\% \improve{2.5}
& 83.6\% $\rightarrow$ 86.5\% \improve{2.9} \\

\quad + deletion
& 82.3\% $\rightarrow$ 83.8\% \improve{1.5}
& 84.5\% $\rightarrow$ 85.9\% \improve{1.4}
& 81.8\% $\rightarrow$ 83.2\% \improve{1.4}
& 83.6\% $\rightarrow$ 85.1\% \improve{1.5} \\

\quad + grouping
& 82.3\% $\rightarrow$ 84.5\% \improve{2.2}
& 84.5\% $\rightarrow$ 86.6\% \improve{2.1}
& 81.8\% $\rightarrow$ 83.9\% \improve{2.1}
& 83.6\% $\rightarrow$ 85.8\% \improve{2.2} \\
\bottomrule
\end{tabularx}

\caption{The indicators of sub-optimization strategies (deletion and grouping) applied to four baselines are shown.}
\label{tab:optimization_results3}
\end{table*}

\subsection{Algorithm of Knowledge Conflicts Between SFT Data}

\begin{algorithm}[t]
\caption{Optimization Strategy Based on Conflict Detection and Conflict Sample Bucketing}
\label{alg:conflict_bucketing}
\begin{algorithmic}[1]

\State \textbf{Input:} 
SFT dataset set $\{D_1, D_2, \dots, D_n\}$; 
semantic similarity threshold $X$; 
number of buckets $B$

\State \textbf{Output:} Optimized SFT dataset

\vspace{0.3em}
\State Initialize conflict group set $\mathcal{C} \leftarrow \emptyset$

\ForAll{sample pair $(s_i, s_j)$ in the dataset}
    \State Compute semantic similarity $\mathrm{Sim}(s_i, s_j)$
    \If{$\mathrm{Sim}(s_i, s_j) > X$}
        \State Use GPT to determine the correctness of $s_i$ and $s_j$
        \If{$s_i$ is incorrect}
            \State Remove $s_i$ from the dataset
        \ElsIf{$s_j$ is incorrect}
            \State Remove $s_j$ from the dataset
        \Else
            \State $\mathcal{C} \leftarrow \mathcal{C} \cup \{(s_i, s_j)\}$
        \EndIf
    \EndIf
\EndFor

\ForAll{conflict group $G \in \mathcal{C}$}
    \State Evenly distribute samples in $G$ into $B$ buckets
\EndFor

\State \Return the optimized SFT dataset

\end{algorithmic}
\end{algorithm}

The "Optimization Strategy Based on Conflict Detection and Conflict Sample Bucketing" method, as illustrated in Algorithm \ref{alg:conflict_bucketing}, initiates by initializing an empty conflict group set. For each pair of samples within the dataset, the semantic similarity is computed by a Sentence-BERT model\cite{edoardo_federici_2022}. If the similarity surpasses a predefined threshold, GPT-4 is employed to assess the correctness of the samples. Incorrect samples are subsequently removed from the dataset, whereas conflicting pairs are incorporated into the conflict group. Following this, the samples within each conflict group are evenly distributed into a designated number of buckets. The process concludes by returning the optimized dataset, ensuring improved quality and reduced conflicts.

\subsection{Deleting and Grouping Conflicting Data}

The experiment is to evaluate the role of knowledge conflict detection and conflict grouping strategies in resolving sample-level contradictions in conflict datasets during  SFT as depicted in Table \ref{tab:optimization_results3}.
In comparison to directly merging datasets, the proposed strategies have consistently improved the accuracy of all baseline models. This outcome demonstrates that integrating conflict detection with perceptually coherent grouping can effectively mitigate the interference caused by conflicting knowledge during batch training. Furthermore, after the removal of conflict data, the accuracy of the model trained with conflict detection and grouping has increased by 1-5\%. This indicates that segregating conflict samples can prevent performance degradation due to label inconsistencies.
Lastly, regarding dynamic grouping, periodically re-evaluating data conflicts (dynamic grouping) ensures superior learning outcomes. By isolating contradictory examples into distinct groups, knowledge conflicts between datasets can be effectively managed. This strategy maximally reduces interference while preserving the value of high-quality samples.

\section{Left-side Forgetting Supplement}
\label{left_side_forgetting}
\subsection{Experimental Datasets}
\begin{table}[ht]
\centering
\renewcommand{\arraystretch}{1.0} 
\setlength{\tabcolsep}{1pt} 
\scalebox{0.9}{
\begin{tabular}{l l c c}
\toprule
\textbf{Dataset} & \textbf{Field} & \textbf{Data Size} & \textbf{Evaluation} \\
\midrule
CNN/DailyMail & News & 312,000 & Acc \\
XSum & Extreme & 226,000 & Acc \\
OpenWebText & Language & 800,000 & Acc \\
\bottomrule
\end{tabular}
}
\caption{\label{s4}SFT Dataset statistics and evaluation index.}
\end{table}

As shown in Table \ref{s4}, the experiment employs datasets from various domains to validate the consistency and effectiveness of the method in enhancing knowledge across different fields, including:

\noindent \textbf{CNN/DailyMail:}\cite{see-etal-2017-get} A widely-used dataset for news summarization tasks, comprising news articles paired with their summaries. This dataset is designed to evaluate the model's ability to generate concise and informative summaries.

\noindent \textbf{XSum:}\cite{Narayan2018DontGM} An extreme summarization dataset, where each sample consists of a news article and a single-sentence summary. This dataset tests the model's capability to produce highly abstractive summaries.

\noindent \textbf{OpenWebText:}\cite{Gokaslan2019OpenWeb} An open-source text dataset derived from Reddit submissions, utilized for training and evaluating language models on diverse and conversational text data.

\paragraph{Parameter Settings for Dynamic Resampling.}
Our dynamic resampling (Algorithm~\ref{alg:dynamic_resampling}) uses two key parameters: evaluation frequency $K$ and performance drop threshold $T_{drop}$. We set $K$ to 500 training steps, balancing timely forgetting detection with computational cost. $T_{drop}$ was empirically set to a 5\% relative performance decrease on a development set, aiming to capture significant degradation while avoiding noise-induced over-triggering. These parameters were based on preliminary experiments; comprehensive sensitivity analysis remains future work.

\begin{algorithm}[t]
\caption{Dynamic resampling}
\label{alg:dynamic_resampling}
\begin{algorithmic}[1]

\State \textbf{Input:} SFT dataset set $\{D_1, D_2, \dots, D_n\}$, training step interval $K$, accuracy drop threshold $T$
\State \textbf{Output:} Optimized SFT model

\vspace{0.3em}
\State Initialize training steps $t = 0$
\State Randomly shuffle all dataset samples

\While{Training is not completed}
    \State Perform $K$ steps of training
    \State Update training steps $t = t + K$

    \For{Each SFT dataset $D_i$}
        \State Calculate current accuracy $A_i(t)$
        \State Calculate accuracy change $\Delta A_i(t) = A_i(t - K) - A_i(t)$

        \If{$\Delta A_i(t) > T$}
            \State Resample from $D_i$ and add to the current training batch
        \EndIf
    \EndFor
\EndWhile

\State \Return the optimized model

\end{algorithmic}
\end{algorithm}

\subsection{Algorithm of Left-side Forgetting}
The "Dynamic Resampling" method, as outlined in Algorithm \ref{alg:dynamic_resampling}, is designed to enhance the performance of an SFT model by adaptively adjusting the data based on accuracy changes. The process begins by initializing training steps and shuffling all datasets. During training, the algorithm performs a fixed number of training steps and updates the step count. For each SFT dataset, it calculates the current accuracy and the change in accuracy compared to the previous interval. If the accuracy drop exceeds the threshold, the algorithm resamples from the corresponding dataset and incorporates these samples into the current training batch.

\begin{table*}
\centering
\begin{tabular}{llr}
\toprule
\textbf{Type} & \textbf{Description} & \textbf{Proportion}  \\
\midrule
I. Base Model Knowledge Limitations & $\mathcal{P}_\text{exist}(x)=0$ generate wrong predict & 18.7\%  \\
II. Conflicts Between SFT and Base Model & $D_\text{JS} > 0.3$ cognitive conflict & 13.2\% \\
III. Knowledge Conflicts Between SFT Data & Wrong answer or multiple positions & 14.1\% \\
IV. Left-side Forgetting  & Previous training data is forgotten & 17.4\% \\
V. Insufficient Training  & SFT data is not fully learned & 14.6\%\\
\bottomrule
\end{tabular}
\caption{\label{tab:taxonomy2}Classification of unlearned phenomena in SFT and their corresponding proportions.}
\end{table*}

\begin{table}[h!]
\centering
\renewcommand{\arraystretch}{1.0} 
\setlength{\tabcolsep}{7.0pt} 
\scalebox{0.9}{
\begin{tabular}{lcccc}
\toprule
\textbf{Position} & \textbf{ROUGE-L} & \textbf{Strategy} & \textbf{Gain} \\
\midrule
First 10\% data & 0.41 & 0.53 & \textcolor{red}{+29\%} \\
Middle data & 0.57 & 0.59 & \textcolor{red}{+3.5\%} \\
Last 10\% data & 0.61 & 0.60 & \textcolor{red}{-1.6\%} \\
\bottomrule
\end{tabular}}
\caption{ROUGE-L results and gain comparison.}
\label{tab:rouge_results4}
\end{table}

\subsection{Analysis of Alleviating Left-sided Forgetting}

The dynamic re-sampling mechanism has significantly mitigated the problem of early - stage data forgetting. The ROUGE - L score of the first 10\% of the training data has increased by 29\% (from 0.41 to 0.53), while the performance of subsequent data has not been significantly impaired (the last 10\% of the data has only decreased by 1.6\%). As shown in Table \ref{tab:rouge_results4}, the gain for data in the middle stage is relatively small (+3.5\%), confirming that the forgetting phenomenon is most pronounced during the initial stage of training.

\begin{table}[ht]
\centering
\renewcommand{\arraystretch}{1.3} 
\setlength{\tabcolsep}{3.6pt} 
\scalebox{0.9}{
\begin{tabular}{l l c c}
\toprule
\textbf{Dataset} & \textbf{Field} & \textbf{Data Size} & \textbf{Evaluation} \\
\midrule
\textbf{AG News} & News & 120,000 & Acc \\
\textbf{IMDB} & Movie Reviews & 50,000 & Acc \\
\textbf{MultiNLI} & NLI & 433,000 & Acc \\
\textbf{QQP} & Quora & 404,000 & Acc \\
\bottomrule
\end{tabular}
}
\caption{\label{s5}SFT Dataset statistics and evaluation index.}
\end{table}

\section{Insufficient Training Supplement}
\label{insufficient}
\subsection{Experimental Datasets}

As shown in Table \ref{s5}, the experiment employs datasets from various domains to validate the consistency and effectiveness of the method in enhancing knowledge across different fields. Specifically, the datasets include:

\noindent \textbf{AG News\cite{Zhang2015CharacterlevelCN}:} The news articles are categorized into four classes: World, Sports, Business, and Sci/Tech. It is commonly used for text classification tasks, evaluating the model's ability to categorize news articles accurately.

\noindent \textbf{IMDB\cite{maas-EtAl:2011:ACL-HLT2011}:} A dataset of movie reviews labeled as positive or negative, widely used for sentiment analysis tasks. This dataset tests the model's capability to understand and classify the sentiment expressed in text.

\noindent \textbf{MultiNLI\cite{N18-1101}:} A dataset for natural language inference (NLI) tasks, containing sentence pairs labeled with their relationship (entailment, contradiction, or neutral). It evaluates the model's ability to understand the logical relationship between two sentences.

\noindent \textbf{Quora Question Pairs\footnote{https://www.kaggle.com/datasets/quora/question-pairs-dataset} :} A dataset consisting of question pairs from Quora, labeled as either duplicate or non-duplicate. It is used for duplicate question detection tasks, assessing the model's ability to identify semantically similar questions.

\subsection{Algorithm of Insufficient Training}

\begin{algorithm}[t]
\caption{Epoch Increment Strategy}
\label{alg:progressive_epoch}
\begin{algorithmic}[1]

\State \textbf{Input:} SFT dataset $D$, initial epoch $E = 1$, evaluation function $\text{Eval}(\cdot)$
\State \textbf{Output:} optimal training round $E_{\text{optimal}}$

\vspace{0.3em}
\State Initialize $P_{\text{best}} = 0$

\While{$P_E \geq P_{\text{best}}$}
    \State Train the model to round $E$
    \State Calculate performance using validation set $P_E = \text{Eval}(\text{Model})$

    \If{$P_E > P_{\text{best}}$}
        \State Update $P_{\text{best}} = P_E$
        \State Increase training round $E = E + 1$
    \Else
        \State Stop training
    \EndIf
\EndWhile

\State \Return the best training round $E_{\text{optimal}} = E - 1$

\end{algorithmic}
\end{algorithm}
The "Epoch Increment Strategy" as illustrated in Algorithm\ref{alg:progressive_epoch}, is designed to identify the optimal number of training epochs for a model by progressively increasing the epoch count and evaluating performance. The strategy commences with an initial epoch count, iteratively trains the model, and assesses its performance on a validation set. If performance improves, the epoch count is incremented, and training continues. Conversely, if no further improvement is detected, training is terminated, and the optimal epoch count is recorded. This approach ensures  the model is trained to achieve the best possible performance without overfitting.

\section{Experiment and Analysis with Olmo2-7B}
\label{olmo}
To further investigate the Incomplete Learning Phenomenon (ILP) and validate our proposed CPT strategies on a recent open-source model, we conducted a series of experiments using OLMo2-7B \cite{olmo20242olmo2furious}.
OLMo2-7B is a 7-billion parameter model, part of the OLMo suite, trained on the Dolma dataset, a 5 trillion token open corpus.
\subsection{Experimental Setup}
\paragraph{Expanded Evaluation Framework.} For a comprehensive assessment of OLMo2-7B's capabilities before and after CPT and SFT, we employed an expanded evaluation framework. This framework assesses performance across four key dimensions, utilizing the following standard benchmarks and their respective metrics:
\begin{itemize}
\item \textbf{General Ability:} MMLU (Massive Multitask Language Understanding) \cite{hendryckstest2021, hendrycks2021ethics} and AGIEval \cite{zhong2023agieval}.
\item \textbf{Reasoning Ability:} BBH (Big-Bench Hard, specifically the 3-shot version) \cite{suzgun2022challenging}.
\item \textbf{Professional Knowledge:} GPQA (Graduate-Level Google-Proof Q\&A Benchmark) \cite{rein2024gpqa} and NQ (Natural Questions) \cite{kwiatkowski-etal-2019-natural}.
\item \textbf{Multilingual Ability:} MMLU-Multi (a multilingual version of MMLU).
\end{itemize}
Performance was measured using the primary accuracy metric reported for each benchmark.

\subsection{Analysis of SFT Data in Relation to OLMo2 Pre-training Corpus}
\label{app:olmo_sft_pretrain_analysis}

To quantitatively understand the extent of pre-training knowledge limitations and potential conflicts OLMo2 might face when fine-tuned on typical Supervised Fine-Tuning (SFT) datasets, we conducted an in-depth analysis comparing our SFT data collections against OLMo2's pre-training corpus (Dolma).

\paragraph{Methodology for Knowledge Relationship Assessment.}
Our methodology involved three primary steps to assess the relationship between individual SFT knowledge items (derived from various SFT datasets used in our study) and the Dolma corpus:

\begin{table*}[h!]
\centering
\renewcommand{\arraystretch}{1.0} 
\begin{tabular}{lccc}
\toprule
\textbf{Dataset Type (SFT)} & \textbf{Total SFT Entries} &  \textbf{Non-Existence Rate (\%)} & \textbf{Conflict Rate (\%)} \\
\midrule
General Ability        & 10,000                 & 18.6                             & 13.8                          \\
Reasoning Knowledge    & 8,000                  & 13.1                              & 10.3                          \\
Professional Knowledge & 9,500                  & 27.4                             & 18.4                          \\
Multilingual Knowledge & 7,500                  & 16.5                             & 15.0                          \\
\midrule
\textbf{Total}           & \textbf{35,000}        & \textbf{19.3}          & \textbf{14.5}                 \\
\bottomrule
\end{tabular}
\caption{\label{tab:olmo_pretrain}Analysis of Knowledge Existence and Conflict between SFT Data and OLMo2 Pre-training Data.}
\end{table*}

\begin{enumerate}
    \item \textbf{Relevant Pre-training Data Retrieval:} Given the 5 trillion token scale of the Dolma corpus, an exhaustive comparison is infeasible. Therefore, for each SFT dataset, we first identified thematic keywords and concepts. We then utilized a distributed indexing cluster built on Elasticsearch\footnote{Elasticsearch BV. Elasticsearch. \url{https://www.elastic.co/elasticsearch/}.} to retrieve the Top-100,000 text snippets from Dolma that were most thematically relevant to these SFT dataset concepts. This step aimed to narrow down the search space to potentially pertinent pre-training data.

    \item \textbf{Precise Semantic Matching:} From these retrieved 100k snippets, we employed an Apache Spark\footnote{Apache Software Foundation. Apache Spark. \url{https://spark.apache.org/}.} cluster in conjunction with a Sentence-BERT model \cite{edoardo_federici_2022}
    to perform fine-grained semantic matching. For each specific knowledge item or query from our SFT samples, this step extracted text segments from the retrieved Dolma snippets that exhibited high semantic similarity to the SFT item.

    \item \textbf{Knowledge Existence and Conflict Evaluation using GPT:} The core assessment was performed using GPT  as an expert evaluator. For each SFT knowledge item, alongside its semantically matched pre-trained text segments from Dolma, GPT was prompted to determine two aspects:
    \begin{itemize}
        \item[(a)] \textbf{Knowledge Existence:} Whether corresponding or semantically equivalent knowledge to the SFT item was present in the provided Dolma segments.
        \item[(b)] \textbf{Knowledge Conflict:} If such knowledge was found, whether the information in the Dolma segments conflicted with the SFT item (e.g., factual discrepancies, outdated information, or contradictory statements).
    \end{itemize}
    The prompt for GPT involved presenting both the SFT item and the retrieved pre-trained snippets, requesting a categorical judgment (exists/not\_exists) along with a brief justification. Based on GPT's judgments, we calculated the \textbf{``Knowledge Non-Existence Rate''} (the proportion of SFT items not found in the relevant retrieved Dolma segments) and the \textbf{``Knowledge Conflict Rate''} (the proportion of SFT items that were found but assessed as conflicting with the Dolma segments).
\end{enumerate}

\paragraph{Statistical Results.}
We applied this analysis pipeline to SFT datasets categorized by the primary capability they aim to instill or evaluate. The aggregated statistical results, showing the Non-Existence Rate and Conflict Rate for different types of SFT data in relation to OLMo2's pre-training corpus, are presented in Table~\ref{tab:olmo_pretrain}.


\paragraph{Discussion of Findings.}
The results in Table~\ref{tab:olmo_pretrain} indicate that a substantial portion of knowledge targeted by common SFT datasets may either be new to OLMo2 or in direct conflict with information encountered during its pre-training (overall Conflict Rate of 14.5\%). For instance, SFT data aimed at "Professional Knowledge" exhibited particularly high rates of both non-existence (27.4\%) and conflict (18.4\%). These figures quantitatively underscore the significant challenges an LLM like OLMo2 faces during SFT, highlighting the necessity for robust mechanisms to inject new knowledge and resolve conflicts, which our CPT strategy aims to provide.

\subsection{CPT Performance and Knowledge Relationship Analysis on OLMo2-7B}
\label{app:olmo_cpt_performance_analysis}

Following the analysis of knowledge gaps and conflicts, we applied our Continued Pre-Training (CPT) strategy to the OLMo2-7B model. The CPT data was specifically curated to address the identified areas of knowledge non-existence and to help resolve conflicts observed between SFT data and OLMo2's pre-training corpus. 

\paragraph{Quantitative Results.}
Table \ref{tab:olmo_cpt_performance} presents the percentage change ($\Delta\%$) in OLMo2-7B's performance on various standard benchmarks post-CPT. These changes are juxtaposed with the "Knowledge Non-Existence Rate" and "Knowledge Conflict Rate" of the SFT data collections used to target each evaluation dimension.

\begin{table*}[htbp]
\centering
\renewcommand{\arraystretch}{0.9}
\scalebox{0.88}{
\setlength{\tabcolsep}{2.8mm}{
\begin{tabular}{llccc} 
\toprule
\textbf{Evaluation Dimension} & \textbf{Benchmark} & \textbf{Non-Existence Rate} & \textbf{Conflict Rate} & \textbf{Performance Change} \\
& & \textbf{(SFT Data, \%)} & \textbf{(SFT Data, \%)} & \textbf{after CPT ($\Delta\%$)} \\
\midrule
General Ability & MMLU & 12.7 & 8.3 & -3.5 \\
General Ability & AGIEval & 15.2 & 6.8 & -2.9 \\
Reasoning Ability & BBH & 9.4 & 4.1 & -1.2 \\
Professional Knowledge & GPQA & 21.5 & 11.6 & -6.8 \\
Professional Knowledge & NQ & 18.3 & 9.7 & -4.3 \\
Multilingual Ability & MMLU-Multi & 23.1 & 15.4 & -8.1 \\
\bottomrule
\end{tabular}}}
\caption{\label{tab:olmo_cpt_performance}CPT Performance Change on OLMo2-7B Across Evaluation Dimensions}
\end{table*}

\paragraph{Discussion of CPT Impact on OLMo2-7B Generalization.}
As demonstrated by Table \ref{tab:olmo_cpt_performance}, the application of CPT on OLMo2-7B led to a decrease in performance across all listed general ability, reasoning, professional knowledge, and multilingual benchmarks. This outcome, while seemingly counterintuitive when CPT is intended for knowledge enhancement, requires careful interpretation in the context of the ILP.

We hypothesize that this observed performance degradation on broad generalization benchmarks reflects the significant cognitive effort and internal recalibration the model undergoes when attempting to integrate substantial amounts of new knowledge and reconcile information that conflicts with its pre-trained biases. The Dolma pre-training corpus is vast, and the knowledge targeted by CPT, while relevant to specific SFT tasks, might represent a relatively small yet potentially disruptive portion compared to the model's overall representations.

Particularly in dimensions like "Professional Knowledge" and "Multilingual Ability," where the SFT data exhibited high Non-Existence and Conflict Rates (up to 23.1\% and 15.4\% respectively, as per Table \ref{tab:olmo_pretrain}), the more pronounced performance drops (e.g., -6.8\% on GPQA, -8.1\% on MMLU-Multi) might signify a period of significant representational adjustment. The model is actively working to incorporate information that is either entirely novel or contradicts its established knowledge base. This process could temporarily disrupt performance on tasks that rely on the stability of its previous, broader knowledge representations.

This suggests a potential trade-off: while our CPT strategy can be effective for targeted knowledge injection and resolving specific conflicts at a granular level, the process of assimilating this specialized or corrective information can have complex, and sometimes initially detrimental, impacts on broadly measured generalization capabilities. This is particularly relevant for highly adaptable open-source models like OLMo2, which might be more sensitive to such shifts. These findings highlight the necessity for carefully calibrated CPT strategies and possibly subsequent SFT stages or other alignment techniques to re-harmonize newly acquired specialized knowledge with the model's general abilities. This observation of a nuanced interplay between targeted knowledge enhancement and general capability retention is an important aspect of understanding and addressing the ILP.

\subsection{Case Studies of Knowledge Conflict Resolution in OLMo2-7B}
\label{app:olmo_conflict_case_studies}
While Appendix~\ref{app:olmo_cpt_performance_analysis} discussed CPT's broader impacts on OLMo2-7B's generalization, this section presents qualitative case studies illustrating its effectiveness in resolving specific knowledge conflicts at a granular level, as detailed in Table~\ref{tab:olmo_conflict_resolution_cases}. These examples show how model outputs shifted post-CPT to better align with SFT data.

\paragraph{Timeliness Conflicts.}
When SFT data presented updated facts conflicting with OLMo2's outdated pre-trained knowledge, CPT helped align the model with the newer information. For instance, when queried about a topic with a recently changed status (e.g., "Who is the current US President?", where the SFT data reflects a more recent administration than the base model’s cutoff), the post-CPT OLMo2 model showed an increased tendency to provide the SFT-aligned, more current answer. In contrast, its pre-CPT responses often defaulted to the older knowledge embedded during pre-training, potentially yielding factually outdated outputs. This shift demonstrates CPT’s effectiveness in resolving knowledge conflicts by prioritizing up-to-date supervised signals over stale pre-trained priors.

\paragraph{Disciplinary Controversies or Evolving Terminology.}

When SFT data introduced perspectives on disciplinary controversies or newer terminology that differed from or were entirely absent in the model's pre-training, CPT facilitated the incorporation of these new viewpoints by recalibrating the model’s internal priors. For example, if pre-trained OLMo2 leaned towards an established theory for a scientific question (e.g., "String Theory" for quantum gravity), and the SFT data emphasized an emerging alternative (e.g., "Loop Quantum Gravity"), the post-CPT model not only acknowledged but often leaned toward the SFT-emphasized perspective in its responses. A similar effect was observed for evolving terminology: when SFT data highlighted modern techniques such as "LayerScale" for neural network regularization—over older, more prevalent terms like "Dropout" from the pre-training era—the post-CPT model adapted its lexical and conceptual usage accordingly. This demonstrates CPT’s capacity to update both factual stances and technical vocabulary in alignment with contemporary supervised signals.

\paragraph{Multilingual Ambiguities and Geo-Specificity.}
CPT also demonstrated utility in resolving conflicts arising from multilingual contexts or geo-specific information not well-represented in the primarily English-centric pre-training. For instance, if an SFT query used a Chinese geographical name (e.g., "\begin{CJK*}{UTF8}{gbsn}库珀蒂诺\end{CJK*}" for Cupertino when asking about "Apple Inc. headquarters"), the post-CPT OLMo2 showed improved understanding and response generation within that specific Chinese language context, compared to a pre-CPT tendency to default to English-based processing or an inability to link the Chinese entity correctly.

\paragraph{Cross-Cultural Differences and Regional Legal Nuances.}
Similarly, for knowledge involving cultural nuances or regional legal differences that might conflict with a more "default" or globally prevalent understanding in the pre-training data, CPT helped sensitize the model to SFT-provided specifics. For example, if SFT data provided context on the meaning of a gesture in a specific culture (e.g., a headshake in India signifying affirmation) that differed from a Western interpretation, post-CPT OLMo2 was more likely to reflect this SFT-aligned, culturally specific understanding. Likewise, for regional legal details (e.g., differing age limits for data privacy for minors across jurisdictions like GDPR vs. China), CPT helped the model adjust its responses based on the geographical context emphasized in the SFT data.

\paragraph{Summary of Case Study Observations.}
These qualitative examples from various conflict types consistently demonstrate that CPT can effectively steer OLMo2-7B's responses towards SFT-aligned knowledge in instances of direct conflict. By encouraging the model to update its internal representations or output tendencies for these particular conflicting concepts, CPT serves as a valuable tool for targeted knowledge correction. This granular effectiveness is crucial for tailoring LLMs to specific, nuanced requirements, complementing the broader (and sometimes complex) performance changes observed on general benchmarks, as discussed in Appendix~\ref{app:olmo_cpt_performance_analysis}.


\subsection{Summary and Implications of OLMo2 Experiments}
The experiments conducted with the OLMo2-7B model provide several critical insights into the Incomplete Learning Phenomenon (ILP) and the application of our proposed Continued Pre-Training (CPT) strategies to a recent, open-source LLM.

First, quantitative analysis (Appendix \ref{app:olmo_sft_pretrain_analysis}) confirmed that significant SFT knowledge portions are absent from or conflict with OLMo2's pre-training corpus. This highlights the prevalence of pre-training knowledge limitations and conflicts as ILP root causes, corroborating findings on other architectures.

Second, CPT's application to OLMo2-7B showed nuanced impacts on generalization benchmarks (Appendix \ref{app:olmo_cpt_performance_analysis}). Observed performance decreases post-CPT likely indicate representational adjustments as the model integrates new or contradictory, task-relevant information. This suggests a trade-off between targeted knowledge injection and preserving broad generalization, especially in adaptable open models, potentially requiring further fine-tuning to re-optimize general capabilities after specialized CPT.

Third, despite the complex interplay with broad generalization metrics, qualitative case studies (Appendix \ref{app:olmo_conflict_case_studies}) demonstrated CPT's clear effectiveness at a granular level. In specific instances of knowledge conflict (e.g., timeliness, disciplinary views, cultural nuances), CPT successfully steered OLMo2's responses to align more closely with SFT-provided knowledge, showcasing its utility as a targeted correction mechanism.

In conclusion, the OLMo2 experiments enrich our understanding of ILP by providing a detailed look at an open-source model's interaction with SFT data and CPT. They affirm the challenges posed by knowledge gaps and conflicts and demonstrate that while CPT is a potent tool for addressing these specific issues at a fine-grained level, its broader impact on model capabilities can be complex and warrants careful, context-dependent application and evaluation-especially when updating factual knowledge. These findings reinforce the need for comprehensive diagnostic frameworks and adaptable mitigation strategies in our main work.

\begin{table*}[htbp]
\centering
\small 
\setlength{\tabcolsep}{5pt} 
\scalebox{0.9}{
\begin{tabular}{p{0.18\textwidth}|p{0.2\textwidth}|p{0.2\textwidth}|p{0.2\textwidth}|p{0.2\textwidth}}
\toprule
\textbf{Conflict Type} & \textbf{Example Scenario (Query/Context)} & \textbf{Pre-trained OLMo2 Knowledge Tendency (Illustrative Output/Bias)} & \textbf{SFT Knowledge Version (Target Output/Fact)} & \textbf{OLMo2 Output Tendency (Post-CPT)} \\
\midrule
Timeliness Conflict & Query: "Who is the current US President?" (SFT data updated to 2023 context) & Might output a president reflecting its pre-training data cutoff (e.g., "Donald Trump"). & Specifies the president as per 2023 SFT data (e.g., "Joe Biden"). & Increased tendency to output the SFT-aligned, more current president. \\
\midrule
Disciplinary Controversy & Query: "What is the optimal theoretical path for quantum gravity?" & May favor a historically prominent theory (e.g., "String Theory"). & SFT data emphasizes an emerging perspective (e.g., "Loop Quantum Gravity"). & Output may present a more balanced view, acknowledge multiple perspectives, or lean towards the SFT-emphasized theory. \\
\midrule
Multilingual Ambiguity / Geo-specificity & Query (SFT in Chinese): "\begin{CJK*}{UTF8}{gbsn}苹果公司总部的坐标是什么？\end{CJK*}" (Coordinates of Apple Inc. headquarters?) & Primarily processes based on English name or common knowledge, may struggle with direct Chinese geo-entity. & SFT provides query/context with the Chinese geographical name "\begin{CJK*}{UTF8}{gbsn}库珀蒂诺\end{CJK*}" (Cupertino). & Improved understanding and response generation within the Chinese language context for the query. \\
\midrule
Cross-cultural Differences & Query: "Meaning of a headshake gesture in India." & Default interpretation might be Western-centric (e.g., negation). & SFT provides context for South Asian interpretation (e.g., affirmation or other nuances). & Output demonstrates more context-dependent judgment, aligning with the SFT-provided cultural nuance. \\
\midrule
Terminology Evolution & Query: "Describe methods for neural network regularization." & May primarily list older, well-established methods (e.g., "Dropout"). & SFT introduces or emphasizes newer terminology/methods (e.g., "LayerScale"). & Output incorporates or gives due consideration to newer terminology/methods, possibly alongside established ones. \\
\midrule
Regional Legal Differences & Query: "Age limit for data privacy protection of minors in [Specific Region]." & May default to a widely known regulation (e.g., GDPR: 16 years). & SFT specifies a different age for the particular region mentioned (e.g., China: 14 years). & Adjusts response based on the specific geographical context provided in the SFT data or query. \\
\bottomrule
\end{tabular}}
\caption{\label{tab:olmo_conflict_resolution_cases}Typical Case Analysis of Knowledge Conflict Resolution in OLMo2-7B via CPT.}
\end{table*}





\end{document}